\newif\ifRAL
\RALfalse 	

\newif\ifTR
\TRfalse 	

\newif\ifPrePrint
\PrePrinttrue

\newif\ifDraft
\Draftfalse

\ifRAL
\documentclass[letterpaper, 10 pt, journal, twoside]{IEEEtran}
\else
\documentclass[letterpaper, 10 pt, conference]{ieeeconf}
\fi

\IEEEoverridecommandlockouts

\ifRAL
\else
\fi		

\makeatletter

\let\proof\@undefined
\let\endproof\@undefined
\makeatother

\usepackage{amsmath} 
\usepackage{amssymb}  
\usepackage{amsthm}
\usepackage{amsfonts}
\usepackage{mathtools}

\usepackage{bm}
\providecommand{\bm}{\pmb}

\theoremstyle{definition}

\theoremstyle{remark}

\usepackage{verbatim}

\usepackage{color}
\usepackage{subcaption}
\usepackage{graphicx}
\usepackage{caption}

\usepackage{optidef}

\usepackage{times}

\usepackage{xspace}

\usepackage[ruled,vlined,linesnumbered]{algorithm2e}
\usepackage[noend]{algorithmic}

\graphicspath{{images/}}

\usepackage[us]{datetime}
\usepackage{caption}
\usepackage[usenames,dvipsnames,table]{xcolor}

\usepackage{hyperref} 
\hypersetup{
    colorlinks,
    citecolor=black,
    filecolor=black,
    linkcolor=black,
    urlcolor=black,
    pdfauthor={},
    pdfsubject={},
    pdftitle={}
}

\usepackage{cleveref}

\usepackage{todonotes}

\usepackage{graphicx}

\usepackage{latexsym}
\usepackage{color}

\usepackage{cite}

\usepackage{caption}

\usepackage{tikz}
\usetikzlibrary{shapes,arrows}

\usepackage{tabularx, booktabs}
\newcolumntype{Y}{>{\centering\arraybackslash}X}
\usepackage{multirow}
\usepackage{paralist}

\usepackage{color}

\newcommand{\vect}[1]{\bm{#1}}		
\newcommand{\nR}[1]{\mathbb{R}^{#1}}		
\newcommand{\upperRomannumeral}[1]{\uppercase\expandafter{\romannumeral#1}}	

\ifRAL 
\author{Yash Vyas$^1$, Marco Tognon$^2$
	
	\thanks{Manuscript received: DD,\,MM,\,YY; Revised DD,\,MM,\,YY ; Accepted  DD,\,MM,\,YY.}
	\thanks{This paper was recommended for publication by [Editor] upon evaluation of the Associate Editor and Reviewers' comments. 
	This work was partially funded by ...} 
	
	\thanks{$^1$ Affiliation, {\tt \footnotesize \href{mailto:first.author@xx.xx}{first.author@xx.xx}, \href{mailto:last.author@xx.xx}{first.author@xx.xx}}
	}
	
	\thanks{$^2$ Affiliation {\tt \footnotesize \href{mailto:last.author@xx.xx}{first.author@xx.xx}}
	}

	\thanks{Digital Object Identifier (DOI): see top of this page.}	
}

\else 
\author{Yash Vyas, Mike Allenspach, Christian Lanegger, Roland Siegwart, Marco Tognon
    \thanks{Y. Vyas, M. Allenspach, C. Lanegger, R. Siegwart, M. Tognon are with the Autonomous Systems Lab, ETH Zurich, 8092 Switzerland. {\tt \footnotesize \href{mailto:yavyas@student.ethz.ch}{mailto:yavyas@student.ethz.ch}, \href{mailto:mtognon@ethz.ch}{mtognon@ethz.ch}. }}
	\thanks{This work was supported by ETH Career Seed Grant SEED-04 20-2 (AEROGUIDE)}
}
\fi

\ifRAL 
\title{Title}

\else 
\title{\bf Modelling and Estimation of Human Walking Gait for Physical Human-Robot Interaction}
\fi

\ifPrePrint
\makeatletter
\def\ps@titlepagestyle{
	\def\@oddfoot{}\def\@evenfoot{}
	\def\@oddhead{\textcolor{red}{\sf\footnotesize Preprint version, final version at http://ieeexplore.ieee.org/ \hfill The 1st IEEE AIRPHARO Workshop}}
	\def\@evenhead{\textcolor{red}{\sf\footnotesize  Preprint version, final version at http://ieeexplore.ieee.org/  \hfill The 1st IEEE AIRPHARO Workshop}}%
}%
\makeatother
\makeatletter
\def\ps@headings{
	\def\@oddfoot{\textcolor{red}{\sf\footnotesize  Preprint version, final version at http://ieeexplore.ieee.org/ \hfill \thepage \;\;~\hfill~\hfill The 1st IEEE AIRPHARO Workshop}}\def\@evenfoot{\hfill\thepage\hfill}
	\def\@oddhead{}\def\@evenhead{}%
}%
\makeatother
\fi	

\ifDraft
\makeatletter
\def\ps@titlepagestyle{
	\def\@oddfoot{}\def\@evenfoot{}
	\def\@oddhead{\textcolor{red}{\sf Draft version  \hfill Confidential}}
	\def\@evenhead{\textcolor{red}{\sf  Draft version  \hfill Confidential}}%
}%
\makeatother
\makeatletter
\def\ps@headings{
	\def\@oddfoot{\textcolor{red}{\sf  Draft version  \hfill Confidential}}\def\@evenfoot{\hfill\thepage\hfill}
	\def\@oddhead{}\def\@evenhead{}%
}%
\makeatother
\usepackage{draftwatermark}
\SetWatermarkText{Confidential draft}
\SetWatermarkScale{0.6}
\SetWatermarkLightness{0.9} 

\fi	

\begin{document}

\maketitle

\begin{abstract}
An approach to model and estimate human walking kinematics in real-time for Physical Human-Robot Interaction is presented. The human gait velocity along the forward and vertical direction of motion is modelled according to the Yoyo-model.
%
We designed an Extended Kalman Filter (EKF) algorithm 
 to estimate the frequency, bias and trigonometric state of a biased sinusoidal signal, from which the kinematic parameters of the Yoyo-model can be extracted. 
Quality and robustness of the estimation are improved by opportune filtering based on heuristics.
%
The approach is successfully evaluated on a real dataset of walking humans, including complex trajectories and changing step frequency over time.

\end{abstract}

\ifRAL 
\else 
{} 
\fi

\section{INTRODUCTION}\label{sec:intro}

%
Humans and robots are increasingly performing tasks in shared environments.
Examples of these include human-robot collaboration in lifting loads, or exoskeleton assistance for neurorehabilitation \cite{lasota2017survey, losey2018review}. Often this involves working with ground-based wheeled robots \cite{brescianini2011ins, li2013human}, or robot arms with contact points \cite{pehlivan2015minimal, losey2018review, li2018physical}.

\begin{figure}[t]
    \centering
    \includegraphics[width=0.4\textwidth]{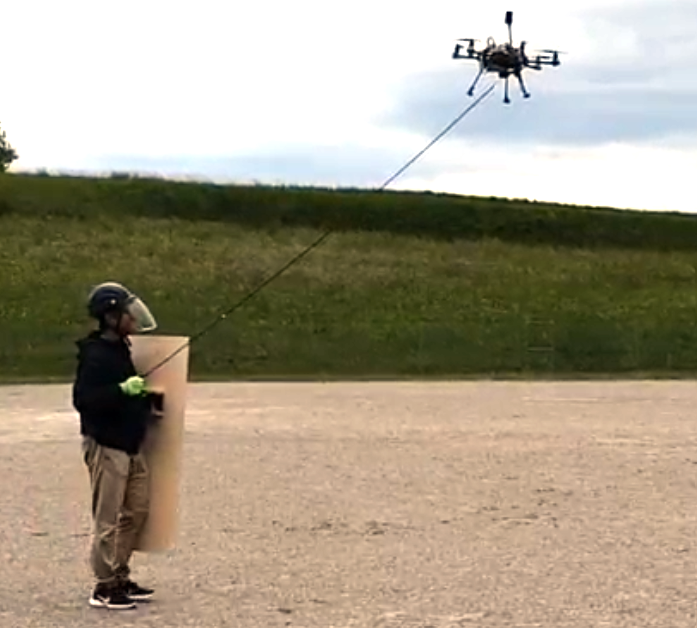}
    \caption{Experiment of human guidance with an aerial robot. The online kinematic estimation can assist the controller in predicting and compensating the oscillations in human motion during walking.}
    \label{fig:experiment}
    \vspace{-1.5em}
\end{figure}

Physical Human-Robot Interaction (PHRI) with flying robots is still a field in its infancy, and mostly involves human avoidance for safety \cite{acharya2017investigation} or compliance to human applied forces \cite{dandrea2013}. Tognon et. al \cite{tognon2021physical} developed a control algorithm for an aerial robot to physically guide a human (e.g., visually impaired) through a tether (see \cref{fig:experiment}). The human and robot are modelled as two simple mass-damper systems connected through a tether. The human position is controlled with a reference force provided by the cable tension, which is proportional to the human position error. 

However, in reality human motion is more complex than what is captured by a mass-damper system, as it contains oscillatory modes arising from cyclic steps during walking. Additionally, the walking gait kinematics vary between persons due to differences in height and stride, and also vary over time with changing walking speeds reflecting human intent and comfort.

To improve human comfort under aerial robot guidance, we seek better adaptation to human motion by incorporating its kinematic model in the control law. This requires modelling the human walking gait and the capability to estimate its parameters online. In particular, we aim at using such estimated model to compensate at the robot level the oscillatory velocity of the human while walking. 
Moreover, a detailed human gait kinematic model and identification of its parameters is relevant for any PHRI application with walking humans. 

From inspecting datasets of the human gait (Section~\ref{ssec:dataset_collection_validation}), we determine the \textit{Yoyo-model}~\cite{carpentier2017centre} to be the most suitable human kinematics model. 
However, the real-time identification of the parameters of such a nonlinear model is still an open problem.
%
The main contribution of this paper is the development of an Extended Kalman Filter (EKF) approach based on \cite{yazdanian2012sinusoidEKF} along with several heuristic filtering steps to estimate the kinematic parameters of the Yoyo-model in real-time (Section \ref{sec:EKF_method}), thereby facilitating incorporation of human velocity-feedback to the control law outlined in \cite{tognon2021physical} as future work.
We validate this approach on a dataset of several walking humans with different trajectories (Section \ref{ssec:dataset_results}), demonstrating that our approach is accurate and responsive to changes in human gait over time.

\section{GAIT MODELLING}\label{sec:gait_modelling}
\begin{figure}[t]
    \centering
    \includegraphics[width=0.5\textwidth]{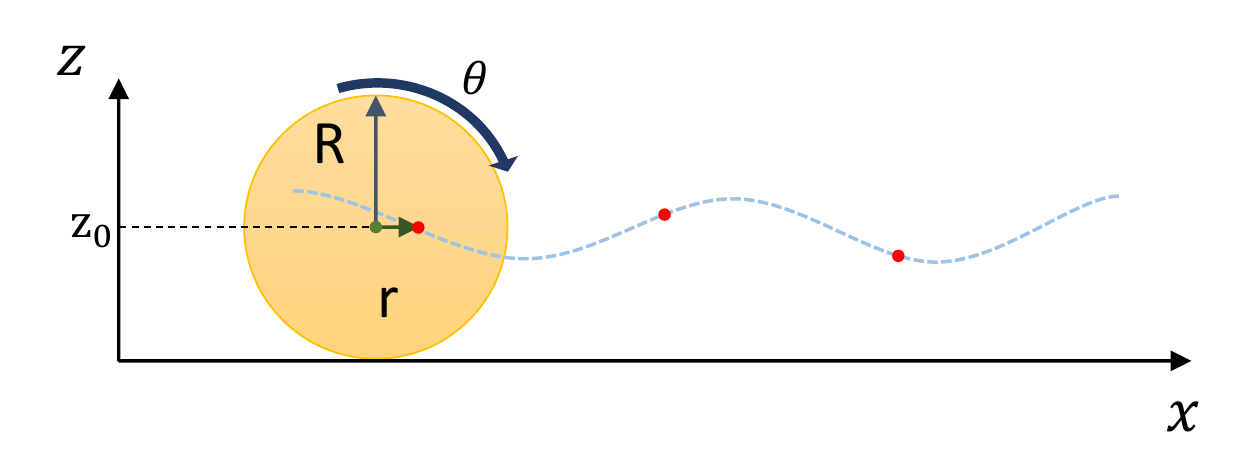}
    \caption{Illustration of the Yoyo-model for the human walking gait.}
    \label{fig:Yoyo_model}
    \vspace{-1.5em}
\end{figure}

The Yoyo-model~\cite{carpentier2017centre} postulates that the position trajectory of the centre of mass of a human while walking can be modelled as a curtate cycloid (see \cref{fig:Yoyo_model}). The primary forward motion is that of a rolling cylinder with radius \(R \in \nR{}_{>0}\). On top of that, forward and vertical motion both contain oscillations corresponding to the respective components of rotation from a smaller inner cylinder with radius \(r \in \nR{}_{>0}\). \(R\) and \(r\) result from the physical attributes of the human (e.g., height) and are assumed to be constant for each human. 
We can write the forward and vertical position of the human, \(x \in \nR{}\) and \(z \in \nR{}\), respectively, in terms of the cylinder angle \(\theta \in \nR{}_{>0}\) (time varying cycloid parameter) and kinematic parameters:
\begin{align}
    x(t) &= R \theta(t) + r \sin(\theta(t)) \label{eq:Yoyo_position_model_x} \\
    z(t) &=  z_0 + r \cos(\theta(t)) \label{eq:Yoyo_position_model_z},
\end{align}
where \(z_0 \in \nR{}_{>0}\) is the z-axis offset of the centre of mass.

In the context of a human being guided along a path by a flying robot, performing parameter estimation in the velocity space is more appropriate \cite{tognon2021physical}.
The angular velocity of the Yoyo-model is defined as \(\omega=\dot{\theta}(t) \in \nR{}\), which corresponds to the step frequency of the human. 
For our application, the human's step frequency can be considered slowly varying and modelled as a constant over several time steps.
We can then write \(\theta(t) = \omega t + \phi\), where \(\phi\) is the initial phase, and \(t \in \nR{} \) is our time variable.
We differentiate (\ref{eq:Yoyo_position_model_x}-\ref{eq:Yoyo_position_model_z}) with respect to time, 
to obtain the differential kinematic relation:
\begin{align}
    v_x(t) &= \dot{x}(t) = 
    R \omega + r \omega \cos(\omega t + \phi), \label{eq:Yoyo_velocity_model_x}\\
    v_z(t) &= \dot{z}(t) =
    - r \omega \sin(\omega t + \phi) \label{eq:Yoyo_velocity_model_z}.
\end{align}
Notice that the human forward velocity is a biased sinusoid, whilst the vertical velocity is unbiased. 

\section{MODEL PARAMETER ESTIMATION}\label{sec:EKF_method}
Although \(R\) and \(r\) are assumed constant, they vary between different humans and are thus included in the estimation problem. 
The step frequency $\omega$ changes according to the human intent (e.g. stopping) and also needs to be estimated.
However, we can assume that 
it is only slowly varying during human walk. 
Hence, the biased sinusoid model (\ref{eq:Yoyo_velocity_model_x}-\ref{eq:Yoyo_velocity_model_z}) is a valid approximation over short time frames
and existing approaches to estimate the bias, amplitude and frequency can be applied.

The methodology we propose is a modified version of the approach presented in \cite{yazdanian2012sinusoidEKF}, which estimates the frequency, bias and amplitude of a slowly-varying sinusoid using an EKF. We define our discrete-time state as the following, where our bias is \(A_0 = R\tilde{\omega}\), amplitude is \(A_1 = r \tilde{\omega}\), and time sample \(k = \frac{t}{T}\) with sampling time \(T\):
\begin{equation}
    \vect{x}(k) =
    \begin{bmatrix}
        x_1(k) \\
        x_2(k) \\
        x_3(k) \\
        x_4(k)
    \end{bmatrix} = 
    \begin{bmatrix}
        A_1 \cos(\Tilde{\omega} k + \phi) \\
        A_1 \sin(\Tilde{\omega} k + \phi) \\
        \Tilde{\omega} \\
        A_0
    \end{bmatrix} .\label{eq:EKF_state}
\end{equation}
Note that \(\Tilde{\omega}\) in (\ref{eq:EKF_state}) refers to the angular velocity estimate in discrete-time with respect to the sampling time,
\(\Tilde{\omega} = \omega T\). We derive our state transition by employing trigonometric identities allowing to write the state at time sample \(k+1\) as a function of the state at time sample \(k\):
\begin{align}
    &\vect{x}(k+1) = 
     \begin{bmatrix}
        x_1(k)\cos(x_3(k)) - x_2(k)\sin(x_3(k)) \\
        x_1(k)\sin(x_3(k)) + x_2(k)\cos(x_3(k))  \\
        x_3(k) \\
        x_4(k)
    \end{bmatrix}. \label{eq:EKF_state_transition}
\end{align}
By using this state formulation, we do not need to estimate the phase \(\phi\), as our states \(x_1(k)\) and \(x_2(k)\) are propagated based on the estimated frequency and current state values. The major modification from the approach used in \cite{yazdanian2012sinusoidEKF} is that the bias \(A_0(k)\) is no longer part of \(x_1(k)\) and \(x_2(k)\). This is possible, since the underlying unbiased sinusoid can directly be observed through the vertical velocity \(v_z(k)\) in (\ref{eq:Yoyo_velocity_model_z}). 
In summary, the measured velocities \(v_x\) and \(v_z\) allow us to observe components of both the unbiased trigonometric state and the bias through the following relationship:
\begin{align}
    v_x(k) &= A_0 + A_1 \cos(\Tilde{\omega} k + \phi) = x_4(k) + x_1(k) \\
    v_z(k) &= - A_1 \sin(\Tilde{\omega} k + \phi) = -x_2(k) .
\end{align}
From this we obtain the following measurement model, where \(H\) is the measurement matrix:
\begin{equation}
    \begin{bmatrix}
        v_x(k) \\
        v_z(k)
    \end{bmatrix} =
    H \vect{x}(k) = 
    \begin{bmatrix}
        1 & 0 & 0 & 1 \\ 0 & -1 & 0 & 0
    \end{bmatrix} \vect{x}(k) . \label{eq:measurement}
\end{equation}
As is the case for the EKF, we assume that state-transition and measurement contain Gaussian zero-mean noise with covariances \(\vect{Q} \in \nR{4 \times 4}\) and \(\vect{V} \in \nR{2 \times 2}\) respectively. The tuning of these noise covariances is essential for the convergence of the EKF state to an accurate estimate of \(\Tilde{\omega}\) and \(A_0\).
Detailed discussions are provided in Section \ref{ssec:dataset_results}. 

Apart from the covariances, we found the sampling frequency is also important for convergence of the estimated parameters away from undesired local minimums such as \(\omega = 0\). One possible reason could be that for very high sampling times, 
the estimator overfits to measurements, ignoring the impact of the state-transition. From trial and error we found that \(\Tilde{\omega} > 0.05\) was optimal for our range of covariance values, with maximum bound \(\Tilde{\omega} < 0.5\) given by the Nyquist-Shannon sampling theorem. In our context, average step frequency is 1-2Hz (6.28-12.6 rad s$^{-1}$), so a sampling time of \(T=0.04s\) was chosen.

The EKF prediction and update equations have been omitted for brevity.
The equations for their implementation using the state-transition (\ref{eq:EKF_state_transition}), measurement (\ref{eq:measurement}) and associated Jacobian matrices outlined here can be found in \cite{yazdanian2012sinusoidEKF}.

In theory, we could extract the parameters for the Yoyo-model directly from our EKF implementation as \(\hat{R}(k) = {x_4(k)}/ {x_3(k)}\), and \(\hat{r}(k) = ({\sqrt{x_1^2(k) + x_2^2(k)}})/{x_3(k)}\). However, applying these equations naively poses several problems.
In the Yoyo-model, our bias \(A_0 = R \omega\) and amplitude \(A_1 = r \omega\) tend to 0 simultaneously as \(\omega \xrightarrow[]{} 0\) when the human stops walking, making the EKF state effectively unobservable. Furthermore, obtaining a tuning that allows for quick response to changes in states \(x_1(k)\) and \(x_2(k)\) while simultaneously exhibiting a slow response to the parameters is not always feasible as the trigonometric state (\(x_1(k)\) and \(x_2(k)\)) and model parameters (\(x_3(k)\) and \(x_4(k)\))  are jointly-estimated in the state-transition (\ref{eq:EKF_state_transition}), and jointly-observed in the measurement model (\ref{eq:measurement}). Due to this, the calculated \(R\) and \(r\) values can vary significantly over time. 

To address these issues, the following heuristic filtering is done on the estimated \(\hat{R}(k)\) and \(\hat{r}(k)\)
:
\begin{itemize}
    \item They are initialised at an approximate value  \(\hat{R}(0)\) and \(\hat{r}(0)\) corresponding to the average human. They are updated only if the estimated parameters \(\hat{\omega}(k) = x_3(k) \) and \(\hat{A}_0 = x_4(k)\) are above particular heuristic values \(\mu_{\omega}\) and \(\mu_{A_0}\), i.e. when the state is fully observable. 
    \item \(\hat{R}(k)\) and \(\hat{r}(k)\) at time sample \(k\) are smoothed by taking the moving average with previous estimates for a defined \(n\) time steps, starting at time sample \(k > n\).
\end{itemize}
The equations for the above are the following:
\small
\begin{align}
    \hat{R}(k) &= 
    \begin{cases}
        \small \frac{1}{n+1}\left(\sum\limits^{k-1}_{i=k-n}\hat{R}(i) + \frac{x_4(k)}{x_3(k)}\right) & 
        \begin{aligned}
            \text{if } & k>n \\
            \text{\& } & x_3(k) > \mu_{\omega} \\
            \text{\& } & x_4(k) > \mu_{A_0}
        \end{aligned}\\
        \hat{R}(k-1) & \text{otherwise},
    \end{cases} \label{eq:R_smoothing_heuristic} \\
    \hat{r}(k) &= 
    \begin{cases}
        \smaller \frac{1}{n+1}\left(\sum\limits^{k-1}_{i=k-n}\hat{r}(i) + \frac{\sqrt{x_1^2(k) + x_2^2(k)}}{x_3(k)}\right) & 
        \begin{aligned}
            \text{if } & k>n, \\
            \text{\& } & x_3(k) > \mu_{\omega} \\
            \text{\& } & x_4(k) > \mu_{A_0}
        \end{aligned}\\
        \hat{r}(k-1) & \text{otherwise}.
    \end{cases} \label{eq:r_smoothing_heuristic}
\end{align}
\normalsize
From this and using the frequency estimate \(\hat{\tilde{\omega}}(k) = x_3(k)\) along with amplitude \(A_1(k) = \sqrt{x_1^2(k) + x_2^2(k)}\), we can reconstruct an estimate for the current velocities:
\begin{align}
    \hat{v}_x(k) &= \hat{R}(k)x_3(k) + \hat{r}(k)x_3(k)\frac{x_1(k)}{A_1(k)} \label{eq:estimation_vx}\\
    \hat{v}_z(k) &= -\hat{r}(k)x_3(k)\frac{x_2(k)}{A_1(k)} \label{eq:estimation_vz} .
\end{align}

\section{EXPERIMENTAL RESULTS}\label{sec:results}
\subsection{Dataset collection and validation of Yoyo-model}
\label{ssec:dataset_collection_validation}
Validation datasets for three different subjects were collected, each performing four different types of walking behaviours:
\begin{enumerate}
    \item Straight line walking at a roughly constant step frequency.
    \item Straight line walking with a varying step frequency, from fast to slow and fast again.
    \item Walking along a roughly rectangular perimeter with smooth turns at roughly a constant step frequency.
    \item Walking in a figure-of-8 pattern at a roughly constant step frequency.
\end{enumerate}
Since obtaining an estimate for the hand velocity is desirable for human guidance with a hand-held tether, the hand position of each subject was tracked using an indoor Motion Capture (MOCAP) system.
Although the original Yoyo-model focuses on the human center of mass, satisfactory results could still be achieved.
This is likely due to the fact that the subjects naturally hold the tether handle in a more-or-less constant body-relative position. 

As the MOCAP system tracks position, Savitzy-Golay filtering was applied 
to obtain velocity measurements.
These values were then resampled at the desired \(25 \rm Hz\) (see Section \ref{sec:EKF_method}) and the forward velocity computed as the 2-norm of the two horizontal components.

Validity of the Yoyo-model was verified through application of a Discrete Fourier Transform (DFT) on the Walk type 1) datasets. 
The expected sinusoidal, \(-90^{\circ}\) degrees phase-offset motion in the forward and vertical directions could be confirmed through this frequency analysis.



\subsection{Application of parameter estimation}
\label{ssec:dataset_results}

\begin{figure}[t!]
    \centering
    \includegraphics[trim={3.2cm 8.5cm 3.3cm 9cm}, clip, width=0.48\textwidth]{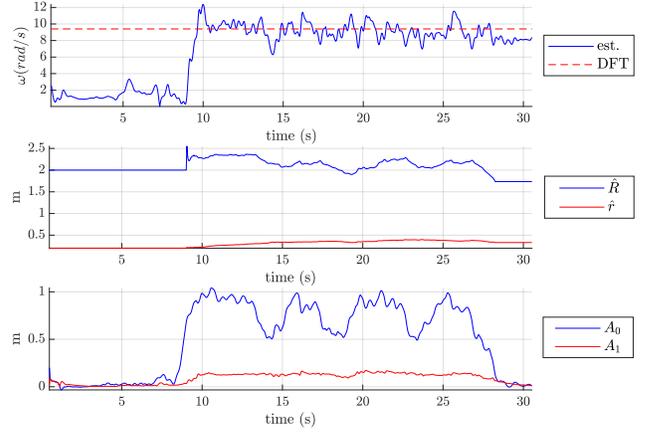}
    \caption{Parameter Estimation for Walk Type 3, Human 1. Top subplot: estimated (est.) normalized absolute value angular velocity estimate \(|\omega(k)|\) compared to the average angular velocity across the data recording obtained via a Discrete Fourier Transform (DFT). Middle subplot: the \(\hat{R}\) and \(\hat{r}\) values under the heuristic filtering and update. Bottom subplot: raw bias \(A_0\) and amplitude \(A_1\) calculated by the EKF.}
    \label{fig:walk3_param_mathias}
\end{figure}

\begin{figure}[t!]
    \centering
    \includegraphics[trim={3.5cm 10cm 3.5cm 10.5cm}, clip, width=0.48\textwidth]{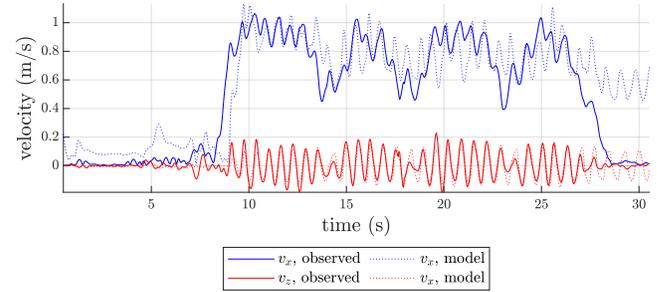}
    \caption{Velocity reconstruction (\ref{eq:estimation_vx}-\ref{eq:estimation_vz}) using the parameters estimated by our model, compared to observed velocities, for Walk Type 3, Human 1.}
    \label{fig:walk3_est_mathias}
    \vspace{-1.5em}
\end{figure}

We tested our algorithm on the datasets described in (Section \ref{ssec:dataset_collection_validation}) using the following covariance values and initializations:

\begin{itemize}
    \item State covariance \(Q = \text{diag}([10^{-5}, 10^{-5}, 10^{-3}, 10^{-3}])\). 
    Note that the relative scaling here is essential for the stable convergence of the parameters over time.
    \item Measurement covariance \(V = \text{diag}([10^{-2}, 10^{-2}])\).
    \item Prior Covariance \(P_0 = \mathbb{I}_{4 \times 4}\), which only has to be relatively large compared to \(Q\).
    \item Initial state \(\vect{x}(0) = [\hat{r}(0) \tilde{\omega}(0),  0, \tilde{\omega}(0), \hat{R}(0) \tilde{\omega}(0)]^T\), where \(\hat{R}(0) = 2 [\rm m]\), \(\hat{r}(0) = 0.2[\rm m]\), and 
    \(\tilde{\omega}(0) = 0.1\) are realistic values for the average human obtained from viewing multiple datasets. Note that the initial values \(x_1(0)\) and \(x_2(0)\) have to be consistent with the phase difference of 90 degrees between sine and cosine.
\end{itemize}

The time length of the heuristic filtering described in (\ref{eq:R_smoothing_heuristic}-\ref{eq:r_smoothing_heuristic}) was set to \(n = 10\), and the corresponding update thresholds were set to \(\mu_{\omega} = 0.1, \mu_{A_0} = 0.1\) through a manual trial-and-error process which balanced smooth parameter changes and speed of convergence to changing parameters.

The implementation of the algorithm in MATLAB took an average of \(54.6 \mu s\) per EKF time step, which proves that it is feasible to be implemented in real-time. Table \ref{tab:dataset_results} summarises the results of the algorithm on each walk type and human test subject, comparing the observed velocities with the estimated velocities (\ref{eq:estimation_vx}-\ref{eq:estimation_vz}). An example implementation for Walk Type 3, Human 1 is shown in \cref{fig:walk3_param_mathias} and \cref{fig:walk3_est_mathias} for reference.

\begin{table}[t]
\centering
\setlength\tabcolsep{2pt}
\begin{tabular}{ | m{1.7cm} | m{2.0cm}| m{2.0cm} | m{2.0cm} |} 
\hline
& \multicolumn{3}{|c|}{\textbf{Mean estimation error (\(10^{-2} (m/s)^{2}\)), }} \\
& \multicolumn{3}{|c|}{\textbf{(\(\pm\) Standard deviation (\(10^{-2} m/s\)))}} \\
\hline
\textbf{Walk Type \& component} & \textbf{Human 1} & \textbf{Human 2}  & \textbf{Human 3} \\ 
\hline
1: \(v_x\) error             & 24.4 (\(\pm\)38.0) & 23.8 (\(\pm\)61.9) & 92.5 (\(\pm\)139) \\
\hspace{0.8em} \(v_z\) error & 0.92 (\(\pm\)1.32) & 2.50 (\(\pm\)5.16) & 2.95 (\(\pm\)4.98) \\
\hline
2: \(v_x\) error             & 216 (\(\pm\)274)   & 237 (\(\pm\)349)   & 184 (\(\pm\)253) \\
\hspace{0.8em} \(v_z\) error & 5.37 (\(\pm\)8.61) & 8.28 (\(\pm\)12.8) & 4.47 (\(\pm\)7.91) \\
\hline
3: \(v_x\)  error            & 32.4 (\(\pm\)46.3) & 63.6 (\(\pm\)116)  & 88.6 (\(\pm\)143) \\
\hspace{0.8em} \(v_z\) error & 1.84 (\(\pm\)3.76) & 4.10 (\(\pm\)7.82) & 4.51 (\(\pm\)8.38) \\
\hline
4: \(v_x\) error             & 29.3 (\(\pm\)70.7) & 76.5 (\(\pm\)164)  & 58.9 (\(\pm\)85.4) \\
\hspace{0.8em} \(v_z\) error & 1.69 (\(\pm\)3.72) & 5.00 (\(\pm\)10.6) & 2.67 (\(\pm\)4.34) \\
\hline
\end{tabular}
\caption{Mean and standard deviation (in brackets) of the squared error for estimation using our algorithm vs. the measured velocity, given with a factor of \(10^{-2}\). The walk type is as specified in Section \ref{ssec:dataset_collection_validation}, and the values for forward (\(v_x\)) and vertical (\(v_z\)) velocities are given separately for comparison.}
\label{tab:dataset_results}
\vspace{-1.5em}
\end{table}

The mean squared estimation error is mainly driven by accuracy in estimation of \(\Tilde{\omega}\), which varies depending on the walking characteristics of human and the trajectory the human is tracking. A high mean error indicates that for most of the human walk  the EKF does not converge to an accurate \(\Tilde{\omega}\), whereas a high standard deviation indicates that the \(\Tilde{\omega}\) varies between accurate and inaccurate throughout the human walk. When the the human is walking at a consistent step frequency for a prolonged period of time, the estimator is more likely to accurately predict the correct \(\Tilde{\omega}\), which improves the overall accuracy of estimation. This can be observed on Walk Type 2, which has a relatively fast time variance in step frequency, and shows higher error on average compared to the other walk types.

The estimation is consistently more accurate for vertical velocity \(v_z(t)\) compared to forward velocity \(v_x(t)\), as \(v_z(t)\) is directly observable in our model, whereas accurate estimation of \(v_x(t)\) requires decoupling of \(x_1(t)\) and \(x_4(t)\) estimation.

It is important to note that a velocity estimation based purely on the EKF state, rather than using \(\hat{R}\) and \(\hat{r}\) would result in a consistently low mean squared estimation error. However, this is an overfit model due to the aforementioned decoupling issues of the estimator. 
Instead, our smoothed parameters (\ref{eq:R_smoothing_heuristic}-\ref{eq:r_smoothing_heuristic}) vary slowly in comparison to the velocity estimate \(\tilde{\omega}\) 
and are thus better at separating the oscillatory modes from the low-frequency forward human velocity.

In many cases where the EKF is initialised with the human being stationary, \(\Tilde{\omega}(k)\) converges to the negative of the true step frequency \(-\Tilde{\omega}(k)\) instead. In \cref{fig:walk3_param_mathias} the absolute value \(|\omega(k)|\) is shown for clarity. As there are two stable equilibriums for the estimator, where the states \(x_1(k)\) and \(x_2(k)\) along with the frequency \(\Tilde{\omega}(k)\) are all negative of their actual values, resulting in the same observed sinusoid, it is likely that once the human starts walking from an unobservable state the EKF can converge to either state based on its observations.

\section{CONCLUSION}\label{sec:conclusion}
In this research we present a novel approach to estimate the state and parameter of a walking human based on the Yoyo-model. An EKF is designed to find the frequency, bias and state of the sinusoidal walking pattern. Due to the observability issues arising from human walk behaviour, we adapt a heuristic filtering algorithm for the kinematic parameters of the Yoyo-model, corresponding to particular physical characteristics of the human. 
We successfully apply this algorithm on a dataset of humans walking with different trajectories and changing step frequencies. 

This algorithm has many potential uses in predicting velocities of walking humans for Human-Robot interaction scenarios. Future work will focus on how the control law in tethered human guidance can be augmented with the oscillation-removed human walking velocity estimates to facilitate a smooth response to changes in human gait. 

\bibliographystyle{IEEEtran}
\bibliography{./bibAlias,./bibCustom}

\end{document}